\documentclass[conference]{IEEEtran}

\IEEEoverridecommandlockouts
\usepackage{graphicx}
\usepackage{xcolor} % optional, only if you really need it
\usepackage{hyperref}

\usepackage{longtable}
\usepackage{multirow} 
\usepackage{tcolorbox}
\usepackage{enumitem}
\usepackage{arydshln}

\definecolor{blond}{rgb}{0.98, 0.94, 0.75}
\definecolor{lightgreen}{rgb}{0.5, 1.0, 0.83}
\definecolor{pastelyellow}{rgb}{0.99, 0.99, 0.59}
\definecolor{pastelorange}{rgb}{1.0, 0.7, 0.28}
\usepackage{float}
\usepackage{soul}
\usepackage{enumitem}
\usepackage[table,xcdraw]{xcolor}
\usepackage[normalem]{ulem}
\useunder{\uline}{\ul}{}
\usepackage{pifont}

\usepackage{tikz}
\usetikzlibrary{arrows.meta,positioning,fit,calc,shapes.geometric}

\def\tsc#1{\csdef{#1}{\textsc{\lowercase{#1}}\xspace}}
\tsc{WGM}
\tsc{QE}
\tsc{EP}
\tsc{PMS}
\tsc{BEC}
\tsc{DE}

\usepackage{cite}
\usepackage{amsmath,amssymb,amsfonts}
\usepackage{textcomp}
\usepackage{subfigure}
\usepackage{enumerate}
\usepackage[flushleft]{threeparttable}
\usepackage{algorithm}
\usepackage{algorithmic}
\usepackage{float}
\usepackage{diagbox}
\usepackage{booktabs}
\usepackage{adjustbox}
\usepackage{comment}

\usepackage{tabularx}

\makeatletter % changes the catcode of @ to 11
\newcommand{\linebreakand}{%
  \end{@IEEEauthorhalign}
  \hfill\mbox{}\par
  \mbox{}\hfill\begin{@IEEEauthorhalign}
}
\makeatother % changes the catcode of @ back to 12

\def\BibTeX{{\rm B\kern-.05em{\sc i\kern-.025em b}\kern-.08em
    T\kern-.1667em\lower.7ex\hbox{E}\kern-.125emX}}
\begin{document}

% \title{A Roadmap of Developing Open-sourced Tools for Data Quality Evaluation and Improvement in Machine Learning}

\title{LongSumEval: Question-Answering Based Evaluation and Feedback-Driven Refinement for Long Document Summarization 
\\ {\large Invited Paper}
}

% \title{Conference Paper Title*\\

\author{\IEEEauthorblockN{1\textsuperscript{st} Huyen Nguyen}
\IEEEauthorblockA{\textit{Cigna Group} \\
\textit{Evernorth Health Services}\\
Austin, Texas, USA \\
HuyenNguyen5@my.unt.edu}

\and
\IEEEauthorblockN{2\textsuperscript{nd} Haoxuan Zhang}
\IEEEauthorblockA{\textit{dept. of Information Science} \\
\textit{University of North Texas}\\
Denton, Texas, USA \\
haoxuanzhang@my.unt.edu}
\and

\IEEEauthorblockN{3\textsuperscript{rd} Yang Zhang}
\IEEEauthorblockA{\textit{dept. of Data Science} \\
\textit{University of North Texas}\\
Denton, Texas, USA \\
yang.zhang@unt.edu}
\and

\linebreakand % <----- NOTE HERE, breaking after the third one!

\IEEEauthorblockN{4\textsuperscript{th} Junhua Ding}
\IEEEauthorblockA{\textit{dept. of Data Science} \\
\textit{University of North Texas}\\
Denton, Texas, USA \\
junhua.ding@unt.edu}
\and

\IEEEauthorblockN{5\textsuperscript{th} Haihua Chen\IEEEauthorrefmark{1}\thanks{This work was
supported in part by the NSF under Grants \#2225229, \#2601493, and \#2231519. Corresponding author: Haihua Chen (Email: haihua.chen@unt.edu)}}
\IEEEauthorblockA{\textit{dept. of Data Science} \\
\textit{University of North Texas}\\
Denton, Texas, USA \\
haihua.chen@unt.edu}
}

\maketitle

% \begingroup\renewcommand\thefootnote{\textsection}
% \footnotetext{Corresponding author.}
% \endgroup

% show page number
\thispagestyle{plain}
\pagestyle{plain}

\begin{abstract}
Evaluating long document summaries remains the primary bottleneck in summarization research. Existing metrics correlate weakly with human judgments and produce aggregate scores without explaining deficiencies or guiding improvement, preventing effective refinement in applications requiring verifiable accuracy. We introduce \textbf{LongSumEval}, a unified framework bridging evaluation and generation through structured question-answering feedback. The framework operationalizes summary quality as answerability and factual alignment of question-answer pairs, generating interpretable scores and actionable feedback that identifies coverage gaps and factual inconsistencies. This resolves the misalignment where evaluation operates independently of generation objectives. Meta-evaluation of our QA-based evaluation module across seven benchmarks demonstrates substantially stronger agreement with human judgments compared to established metrics. Structured feedback enables significant quality improvements through self-refinement without retraining. By demonstrating that evaluation feedback can serve as executable instructions for generation, this work establishes a generalizable paradigm for aligning assessment with improvement, with direct implications for controllable text generation requiring verifiable accuracy and transparent quality control. All code and datasets will be released in GitHub for reproducibility. 
\end{abstract}

\begin{IEEEkeywords}
 Text summarization, Large language model, Question answering, Self-refinement, Meta-evaluation
\end{IEEEkeywords}

\section{Introduction}
\label{intro}
The exponential growth of textual information overwhelms manual review capacity, necessitating automatic summarization systems that distill lengthy documents into concise representations. Recent advances in large language models (LLMs) have transformed the field~\cite{luo2024comprehensive,zhang2025comprehensive}, enabling deployment in critical domains including biomedical literature review~\cite{koh2022far,croxford2025current}, legal document analysis~\cite{mentzingen2025effectiveness}, and clinical decision support~\cite{hark2024power}, where efficient information access directly impacts decision quality.

Despite significant progress in generation capabilities, evaluating summarization quality remains a fundamental bottleneck~\cite{nguyen2024comparative, zhang2025comprehensive}. Traditional metrics relying on lexical overlap (ROUGE~\cite{chin2004rouge}, BLEU~\cite{papineni2002bleu}) or semantic similarity (BERTScore~\cite{zhangbertscore}) correlate poorly with human judgments on factual consistency, coverage, and fluency~\cite{koh2022far,fabbri2021summeval}. They penalize valid paraphrases, fail without reference summaries~\cite{deutsch2021towards}, and exhibit fundamental misalignment with human quality perceptions. LLM-generated summaries often receive lower ROUGE scores despite superior linguistic quality~\cite{goyal2023news}. Recent evaluations on long documents reveal that current automatic metrics fail to detect unfaithful claims effectively~\cite{kim2024fables}, particularly in high-stakes domains like legal and medical summarization~\cite{ding2023quality, croxford2025current}. Moreover, conventional metrics produce opaque scores without diagnostic feedback~\cite{deutsch2021towards,scialom2021questeval}.

Question-answering (QA) based evaluation offers a promising alternative: high-quality summaries should answer salient questions about source documents while maintaining verifiable claims~\cite{deutsch2021towards,scialom2021questeval,wang2020asking, ding2024evaluation}. Representative methods include QuestEval~\cite{scialom2021questeval} for bidirectional consistency assessment, QAEval~\cite{deutsch2021towards} for importance-weighted coverage evaluation, and QAGS~\cite{wang2020asking} for factual inconsistency detection, with extensions targeting dialogue evaluation~\cite{honovich-etal-2021-q2}, fine-grained assessment~\cite{zhang2025qapyramid}, and answer quality scoring~\cite{durmus-etal-2020-feqa}. However, existing frameworks face critical limitations: they generate predominantly factoid questions while neglecting conceptual information~\cite{durmus-etal-2020-feqa}, return aggregated scores without structured feedback specifying missing or incorrect information~\cite{deutsch2021towards,scialom2021questeval}, and struggle with computational costs and context constraints~\cite{koh2022far}. These challenges intensify for long documents exceeding 10,000 words, which contain distributed information, dense technical content, and complex argumentation structures~\cite{koh2022far,maynez2020faithfulness,cao2018faithful}. Evaluation frameworks designed for short news articles fail to scale effectively~\cite{luo2024comprehensive,zhang2025comprehensive}, struggling to identify whether summaries capture scattered key information or introduce subtle factual distortions~\cite{kryscinski2020evaluating,hemamou2024scaling}.

Beyond evaluation, the disconnect between assessment and improvement represents a critical gap. While self-refinement methods demonstrate that LLMs can iteratively improve outputs with explicit feedback~\cite{madaan2024self,chen2024teaching,kim2024prometheus}, existing approaches face two limitations. Summarization-specific methods like SummIt~\cite{zhang-etal-2023-summit} and multi-fact correction~\cite{dong-etal-2020-multi} lack explicit feedback specifications, while recent advances in structured feedback~\cite{wu-etal-2025-meta,yang2025lighthouse,brugge2024large} operate independently of evaluation metrics, creating misalignment between assessment and correction~\cite{gou2024critic}. For long documents, generic critiques prove insufficient. Targeted refinement requires precise identification of coverage gaps and factual inconsistencies~\cite{koh2022far}.

Therefore, we propose \textbf{LongSumEval}, a unified framework integrating LLM-based QA evaluation with feedback-driven refinement. Our approach introduces two key innovations. First, we design a QA evaluation module assessing coverage and factual consistency~\cite{koh2022far,fabbri2021summeval} through diverse question types~\cite{durmus-etal-2020-feqa}, producing structured feedback that identifies unanswered questions (coverage gaps) and inconsistent fact triplets (erroneous claims with ground-truth corrections). Unlike conventional QA metrics returning scalar scores~\cite{deutsch2021towards,scialom2021questeval,scialom2019answers}, our method enables transparent diagnosis through human-inspectable question-answer pairs. Second, inspired by meta-evaluation and critique-guided improvement~\cite{wu-etal-2025-meta,yang2025lighthouse}, we develop a self-refinement module converting structured feedback into natural language instructions for targeted revision. Unlike previous frameworks lacking explicit feedback specifications~\cite{madaan2024self,zhang-etal-2023-summit}, this approach provides fine-grained, actionable guidance for correcting specific deficiencies.

To validate LongSumEval's effectiveness, we formulate three research questions:
\begin{itemize}
    \item \textbf{RQ1:} How effectively does LLM-based QA evaluation correlate with human quality judgments across diverse domains and document lengths compared to conventional QA-based metrics?
    \item \textbf{RQ2:} What impact do design choices (question generation strategies, answer similarity measures, and LLM model selection) have on evaluation reliability and domain robustness?
    \item \textbf{RQ3:} Does iterative refinement guided by structured QA-based feedback lead to measurable improvements in summary coverage and factual consistency without model fine-tuning?
\end{itemize}

We conduct systematic meta-evaluation across seven human-annotated datasets spanning five domains (news, scientific writing, government reports, social media, and patents) with document lengths ranging from 400 to 27,000 words. For RQ1, we benchmark our QA-based evaluation module against established QA-based metrics including QuestEval~\cite{scialom2021questeval}, QAEval~\cite{deutsch2021towards}, and SummaQA~\cite{scialom2019answers}, demonstrating superior correlation with human judgments particularly on long-document benchmarks. For RQ2, we analyze how question diversity~\cite{durmus-etal-2020-feqa}, similarity thresholds, and LLM backend capabilities influence metric reliability across domains. Addressing RQ3, we demonstrate that structured feedback-driven self-refinement~\cite{madaan2024self} achieves substantial quality improvements, with coverage scores increasing by up to 83.72\% and consistency scores improving by up to 47.42\% on initially low-quality summaries.

Our contributions are fourfold:
\begin{itemize}
    \item We design an interpretable LLM-based QA evaluation framework that produces fine-grained structured feedback alongside quantitative scores, enabling transparent reference-free assessment of long document summaries across coverage and factual consistency dimensions.
    
    \item We introduce a feedback-driven self-refinement mechanism that bridges evaluation and generation by converting structured QA-based feedback into actionable natural language instructions, enabling iterative summary improvement through targeted revisions that address specific coverage gaps and factual inconsistencies.
    
    \item We conduct a comprehensive meta-evaluation of our QA-based evaluation module across seven human-annotated datasets spanning five domains and multiple document lengths, demonstrating superior correlation with human judgments compared to existing QA-based metrics, achieving Kendall's $\tau_b$ up to 0.683 for consistency evaluation and 0.738 for coverage assessment on long-document benchmarks.

    \item We introduce \textbf{PatentSumEval}, a human-annotated benchmark for patent document summarization evaluation comprising 30 patents and 180 system summaries with multi-dimensional quality annotations, and release evaluation code and prompts to facilitate reproducibility and future research.
\end{itemize}

\section{Related Works}

\subsection{Automatic Evaluation of Text Summarization}

Early evaluation metrics for text summarization relied on lexical overlap measures such as ROUGE~\cite{chin2004rouge} and BLEU~\cite{papineni2002bleu}. While computationally efficient, these metrics correlate poorly with human judgments and fail to assess factual correctness~\cite{fabbri2021summeval}. Learned metrics like BERTScore~\cite{zhangbertscore} improved semantic similarity measurement but still require reference summaries and struggle with factual consistency evaluation~\cite{koh2022far}.

Question-answering based evaluation addresses these limitations by operationalizing the intuition that high-quality summaries should answer key questions about source documents. QuestEval~\cite{scialom2021questeval} generates questions from both documents and summaries to assess consistency through answer alignment. QAEval~\cite{deutsch2021towards} introduces importance weighting and evaluates coverage via question answerability. QAGS~\cite{wang2020asking} focuses on detecting factual inconsistencies by verifying summary-derived questions against source documents. Extensions include Q\textsuperscript{2}~\cite{honovich-etal-2021-q2} for dialogue evaluation and QAPyramid~\cite{zhang2025qapyramid} for fine-grained content assessment.

Despite progress, existing QA-based metrics have key limitations. Most focus on factoid questions targeting named entities, overlooking conceptual information in \textit{why} and \textit{how} questions~\cite{durmus-etal-2020-feqa}. They produce numerical scores but lack actionable feedback for improvement. They also struggle with long documents due to context constraints and computational costs~\cite{koh2022far}. Design choices regarding question types, similarity measures, and aggregation remain underexplored.

\subsection{Self-Refinement and Iterative Improvement}

Traditional generation systems operate in a single pass without error correction mechanisms. Self-Refine~\cite{madaan2024self} demonstrated that large language models can iteratively critique and improve their outputs when given appropriate prompting, inspiring extensions to code generation~\cite{chen2024teaching}, instruction following~\cite{kim2024prometheus}, and creative writing~\cite{yang-etal-2022-re3}.

In summary, refinement approaches target factual errors and coverage gaps. Multi-Fact Correction~\cite{dong-etal-2020-multi} uses QA models to identify and replace inconsistent entities. FactCC~\cite{kryscinski2020evaluating} and SummaC~\cite{laban2022summac} detect inconsistencies via natural language inference but lack correction mechanisms. Recent work on retrieval-augmented generation explores iterative refinement with structured feedback. IRAGKR~\cite{du2025iragkr} demonstrates that fine-grained knowledge refinement through iterative retrieval improves output accuracy, while adaptive retrieval methods~\cite{han2025adaptive} dynamically adjust strategies based on generation quality, highlighting the value of feedback-driven adaptation.

A critical gap remains: existing methods lack structured, fine-grained feedback specifying what is missing or incorrect. Generic critiques provide insufficient guidance for targeted corrections in long documents. Moreover, refinement methods typically operate independently of evaluation metrics, creating a disconnect between assessment and improvement. Our LongSumEval framework addresses these gaps by converting structured QA-based evaluation into actionable natural language instructions, enabling targeted refinement that bridges evaluation and generation.

\section{Methodology}

\subsection{Task Definition}

We address the problem of evaluating and improving automatically generated summaries for long documents. Formally, given a source document \(D\) and a generated summary \(S\), we define two interrelated tasks:

\paragraph{Task 1: Multi-dimensional Quality Assessment.}
Design an evaluation function \(\mathcal{E}\) that computes quality scores and structured feedback:
\begin{equation}
\mathcal{E}(D, S) \rightarrow (\text{score}_{\text{cov}}, \text{score}_{\text{cons}}, \mathcal{F}_{\text{cov}}, \mathcal{F}_{\text{cons}})
\end{equation}
where \(\text{score}_{\text{cov}}, \text{score}_{\text{cons}} \in [0, 1]\) represent coverage and factual consistency scores, respectively, and \(\mathcal{F}_{\text{cov}}, \mathcal{F}_{\text{cons}}\) denote structured feedback identifying coverage gaps and factual inconsistencies.

\paragraph{Task 2: Iterative Summary Refinement.}
Design a refinement function \(\mathcal{R}\) that generates improved summaries through iterative feedback-guided refinement:
\begin{equation}
S^{(i+1)} = \mathcal{R}(D, S^{(i)}, \tilde{\mathcal{F}}^{(i)}), \quad i = 0, 1, \ldots, I_{\text{max}}-1
\end{equation}
where \(S^{(i)}\) denotes the summary at iteration \(i\), \(\mathcal{F}^{(i)} = (\mathcal{F}_{\text{cov}}^{(i)}, \mathcal{F}_{\text{cons}}^{(i)})\) represents the structured feedback obtained from \(\mathcal{E}(D, S^{(i)})\), \(\tilde{\mathcal{F}}^{(i)}\) is the natural-language feedback constructed from \(\mathcal{F}^{(i)}\), and \(I_{\text{max}}\) is the maximum number of iterations. The process terminates when quality thresholds are satisfied or the iteration limit is reached.

\subsection{The LongSumEval Framework}

We propose LongSumEval, a comprehensive framework for evaluating and improving long document summaries through question-answering alignment. The framework consists of two interconnected components that address the tasks defined above:

\begin{itemize}
\item \textbf{QA-Based Evaluation Module}: Implements the evaluation function \(\mathcal{E}\) that assesses summaries on coverage and factual consistency, producing both quantitative scores and structured feedback.

\item \textbf{Self-Refinement Module}: Implements the refinement function \(\mathcal{R}\) that leverages evaluation feedback to iteratively improve summary quality.
\end{itemize}

\begin{figure}[t]
\centering
\includegraphics[width=\linewidth]{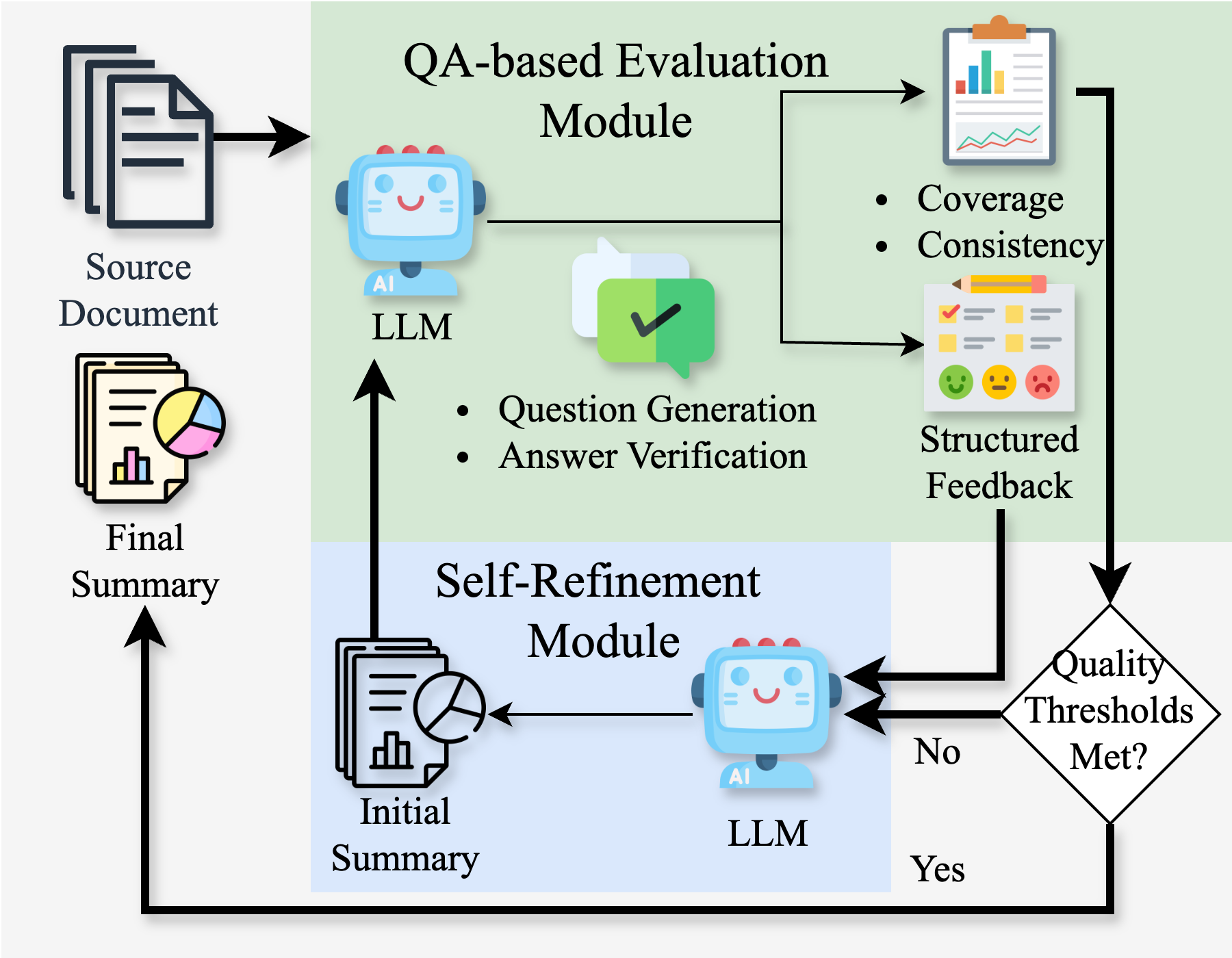}
\caption{Overview of LongSumEval framework. \textbf{Evaluation Module:} Computes coverage and factual consistency scores via LLM-based question answering and produces structured feedback. \textbf{Self-Refinement Module:} Uses the feedback to iteratively revise the summary until quality thresholds are met.
}
\label{fig:longsumeval-framework}
\end{figure}

Both modules address two critical quality dimensions~\cite{fabbri2021summeval, koh2022far}: \textbf{Coverage} (the extent to which the summary captures salient information from the source document) and \textbf{Factual Consistency} (the degree to which factual statements in the summary align with the source document)---the core intuition is that a high-quality summary should answer salient questions about the source (coverage) while keeping its stated facts verifiable against the document (consistency). Figure~\ref{fig:longsumeval-framework} summarizes the overall workflow.

\subsection{QA-Based Evaluation Module}

The evaluation module implements function \(\mathcal{E}\) through question-answering alignment. We denote \(Q_d\) as the set of questions generated from document \(D\), \(Q_s\) as questions from summary \(S\), and \(A_s\), \(A_d\) as corresponding answer sets. The module uses a similarity threshold \(\tau\) to distinguish consistent from inconsistent facts.

Unlike traditional metrics relying on n-gram overlap~\cite{chin2004rouge, papineni2002bleu} or requiring extensive training~\cite{deutsch2021towards,scialom2021questeval}, this module leverages LLM question-answering capabilities to extract and verify factual information at a fine-grained level. This design offers three advantages: (1) \textit{interpretability}—evaluation is grounded in explicit, human-inspectable questions and answers; (2) \textit{flexibility}—adaptation to different domains without retraining; and (3) \textit{actionability}—structured feedback directly indicates missing or inconsistent information. Algorithm~\ref{alg:longsumeval} provides an overview of the evaluation module.

\begin{algorithm}[t]
\centering
\caption{LongSumEval Evaluation Module}
\label{alg:longsumeval}
\begin{algorithmic}[1]
\REQUIRE Source document \(D\), Generated summary \(S\), Similarity threshold \(\tau\)
\ENSURE Coverage score \(\text{score}_{\text{cov}}\), Consistency score \(\text{score}_{\text{cons}}\), Feedback \((\mathcal{F}_{\text{cov}}, \mathcal{F}_{\text{cons}})\)

\STATE \textbf{// Coverage Evaluation}
\STATE \(Q_d \gets \text{GenerateQuestions}(D)\) \COMMENT{Key questions from document}
\STATE \(n_a \gets\) Count answerable questions from \(S\)
\STATE \(\text{score}_{\text{cov}} \gets n_a / |Q_d|\)
\STATE \(\mathcal{F}_{\text{cov}} \gets\) Unanswered questions

\STATE
\STATE \textbf{// Factual Consistency Evaluation}
\STATE \((Q_s, A_s) \gets \text{GenerateQAPairs}(S)\) \COMMENT{Facts from summary}
\STATE \(A_d \gets \text{ExtractAnswers}(D, Q_s)\) \COMMENT{Ground truth from document}
\STATE Compute consistency score:
\STATE \quad \(\text{score}_{\text{cons}} \gets \frac{1}{|Q_s|} \sum_{i=1}^{|Q_s|} \mathbb{I}(\text{Sim}(a_i^s, a_i^d) > \tau) \cdot \text{Sim}(a_i^s, a_i^d)\)
\STATE \(\mathcal{F}_{\text{cons}} \gets\) Inconsistent fact triplets \(\{(q_i, a_i^s, a_i^d) \mid \text{Sim}(a_i^s, a_i^d) \leq \tau\}\)

\STATE
\RETURN \((\text{score}_{\text{cov}}, \text{score}_{\text{cons}}, \mathcal{F}_{\text{cov}}, \mathcal{F}_{\text{cons}})\)
\end{algorithmic}
\end{algorithm}

\subsubsection{Fact Extraction via Question Generation}

The question generation component extracts and represents factual information as question-answer pairs, enabling explicit verification of text content~\cite{deutsch2021towards,scialom2021questeval}. Given a text (document or summary), an LLM generates questions probing key information, extracts corresponding answers, and ranks questions by importance using Prompt~\ref{prompt:qa-generation}.

{\small
\begin{tcolorbox}[
    colback=blue!5, 
    colframe=blue!50!black, 
    boxrule=1pt, 
    rounded corners, 
    title=\textbf{Prompt III-C1: Question-Answer Generation}, 
    fonttitle=\bfseries,
    label={prompt:qa-generation}
]

\textbf{Task} \\
Generate questions that capture key information in the given text, extract corresponding answers, and rank questions by importance.

\vspace{4pt}
\textbf{Requirements}
\begin{itemize}[leftmargin=1.5em, itemsep=2pt, parsep=0pt, topsep=2pt]
    \item Generate \(n\) questions (\(n = 6\text{--}12\) for documents, \(n = 4\text{--}10\) for summaries)
    \item Use diverse question types: What, When, Where, Who, How, Why
    \item Focus on key information, not trivial details
    \item Rank by importance (Rank 1 = most important)
\end{itemize}

\vspace{4pt}
\textbf{Input} \\
Text \(T\): \textit{<text\_content>}

\vspace{4pt}
\textbf{Output Format}
\begin{verbatim}
Question [Rank]: <question_text>
Answer: <answer_text>
...
\end{verbatim}

\end{tcolorbox}
}

To capture both factual details and higher-level concepts, the LLM generates diverse question types including \textit{What}, \textit{When}, \textit{Where}, \textit{How}, and \textit{Why} questions. This extends beyond the named-entity focus of prior QA-based metrics~\cite{deutsch2021towards,scialom2021questeval, wang2020asking}. The importance ranking enables hierarchical evaluation by focusing on top-ranked questions, assessing whether a summary prioritizes critical facts over peripheral details.

\subsubsection{Coverage Assessment}

Coverage assessment quantifies how comprehensively a summary captures key document information~\cite{deutsch2021towards, fabbri2021summeval}. The procedure implements document-to-summary alignment: if a summary adequately covers the source document, it should answer questions about the document's key content. This avoids requiring reference summaries, which are often unavailable for long documents~\cite{koh2022far}.

The implementation follows three steps. First, we generate questions \(Q_d\) from the document representing key information (Prompt~\ref{prompt:qa-generation}). Second, for each question, we extract an answer from the summary using Prompt~\ref{prompt:answer-extraction}; the LLM returns "UNANSWERABLE" if the summary lacks sufficient information. Third, we compute the coverage score as the ratio of answerable questions \(n_a\) to total questions \(|Q_d|\). The score ranges from 0 (no coverage) to 1 (complete coverage).

For fine-grained analysis, we extend the metric to consider only the top-k most important questions. Beyond numerical scores, the evaluation module produces structured feedback \(\mathcal{F}_{\text{cov}}\) containing unanswered questions, explicitly identifying missing key information to guide refinement.

{\small
\begin{tcolorbox}[
    colback=green!5, 
    colframe=green!50!black, 
    boxrule=1pt, 
    rounded corners, 
    title=\textbf{Prompt III-C2: Answer Extraction for Coverage Assessment}, 
    fonttitle=\bfseries,
    label={prompt:answer-extraction}
]

\textbf{Task} \\
Extract answers to the given questions from the summary. If a question cannot be answered, respond with "UNANSWERABLE".

\vspace{4pt}
\textbf{Requirements}
\begin{itemize}[leftmargin=1.5em, itemsep=2pt, parsep=0pt, topsep=2pt]
    \item Extract concise, accurate answers
    \item Base answers only on information explicitly present in the summary
    \item Return "UNANSWERABLE" if information is insufficient or absent
    \item Do not infer or generate information beyond what is stated
\end{itemize}

\vspace{4pt}
\textbf{Input} \\
Summary \(S\): \textit{<summary\_content>} \\
Questions \(Q_d\): \textit{<question\_list>} 

\vspace{4pt}
\textbf{Output Format}
\begin{verbatim}
Question: <question_text>
Answer: <answer_text | "UNANSWERABLE">
...
\end{verbatim}

\end{tcolorbox}
}

\subsubsection{Factual Consistency Assessment}

Factual consistency assessment detects whether a summary introduces claims that contradict the source document. This is critical for neural abstractive models prone to hallucination~\cite{maynez2020faithfulness, cao2018faithful,kryscinski2020evaluating}.

The procedure implements summary-to-document verification: a summary is factually consistent if answers to questions derived from its claims align with those obtained from the source document. This enables fact-level verification rather than holistic scoring~\cite{laban2022summac}.

{\small
\begin{tcolorbox}[
    colback=orange!5, 
    colframe=orange!50!black, 
    boxrule=1pt, 
    rounded corners, 
    title=\textbf{Prompt III-C3: Answer Extraction for Consistency Verification}, 
    fonttitle=\bfseries,
    label={prompt:consistency-verification}
]

\textbf{Task} \\
Extract ground truth answers to the given questions from the source document for verifying summary consistency.

\vspace{4pt}
\textbf{Requirements}
\begin{itemize}[leftmargin=1.5em, itemsep=2pt, parsep=0pt, topsep=2pt]
    \item Extract accurate answers based solely on the document
    \item Provide concise but complete answers
    \item Return "UNANSWERABLE" if information is not present in the document
    \item Do not add information beyond what is explicitly stated
\end{itemize}

\vspace{4pt}
\textbf{Input} \\
Document \(D\): \textit{<document\_content>} \\
Questions \(Q_s\): \textit{<question\_list>}

\vspace{4pt}
\textbf{Output Format}
\begin{verbatim}
Question: <question_text>
Answer: <answer_text | "UNANSWERABLE">
...
\end{verbatim}

\end{tcolorbox}
}

The implementation follows four steps. First, we generate question-answer pairs \((Q_s, A_s)\) from the summary representing factual claims (Prompt~\ref{prompt:qa-generation}). Second, for each question, we extract the ground truth answer from the document using Prompt~\ref{prompt:consistency-verification}, forming answer set \(A_d\). Third, we compute similarity between each answer pair \((a_i^s, a_i^d)\) using one of three measures: \textit{Exact/Partial Match} (string or Jaccard similarity), \textit{ROUGE-1 F1}~\cite{chin2004rouge} (unigram overlap), or \textit{Cosine Similarity} (semantic similarity via embeddings). Fourth, we apply threshold \(\tau\) to distinguish consistent from inconsistent facts, computing the score by averaging similarity scores exceeding the threshold.

The threshold filters low-similarity pairs indicating errors and amplifies quality differences. For fine-grained analysis, we extend the metric to focus on the most important facts. Structured feedback \(\mathcal{F}_{\text{cons}}\) contains inconsistent fact triplets specifying the question, summary answer, and document answer, pinpointing specific errors for correction.

\subsection{Self-Refinement Module}

The self-refinement module implements function \(\mathcal{R}\) to leverage structured feedback from the evaluation module for iterative improvement. Traditional summarization approaches generate summaries in a single pass without identifying and correcting specific deficiencies. This module addresses this by establishing a feedback loop where evaluation results directly inform targeted refinement~\cite{madaan2024self}. 

The iterative procedure, illustrated in Figure~\ref{fig:longsumeval-framework}, operates as follows. Given a source document, the system generates an initial summary \(S^{(0)}\), then iteratively evaluates and refines it. At each iteration, the evaluation module produces quality scores and structured feedback identifying specific deficiencies. Based on quality thresholds \(T_{\text{cov}}\) and \(T_{\text{cons}}\), the system determines which dimension requires improvement. The key innovation lies in converting structured feedback into natural language instructions that explicitly identify missing information or inconsistent facts, enabling targeted corrections without requiring the LLM to independently diagnose problems in lengthy documents. Algorithm~\ref{alg:self-refinement} summarizes the self-refinement procedure.

\begin{algorithm}[t]
\centering
\caption{LongSumEval Self-Refinement Module. \(\tilde{\mathcal{F}}^{(i)}\) denotes natural-language feedback constructed from structured evaluation feedback \(\mathcal{F}^{(i)}\).}
\label{alg:self-refinement}
\begin{algorithmic}[1]
\REQUIRE Source document \(D\), Quality thresholds \(T_{\text{cov}}, T_{\text{cons}}\), Maximum iterations \(I_{\text{max}}\)
\ENSURE Refined summary \(S^*\)

\STATE \(S^{(0)} \gets \text{InitialSummarize}(D)\) \COMMENT{Generate initial summary}
\STATE \(i \gets 0\)

\WHILE{\(i < I_{\text{max}}\)}
    \STATE \((\text{score}_{\text{cov}}, \text{score}_{\text{cons}}, \mathcal{F}_{\text{cov}}^{(i)}, \mathcal{F}_{\text{cons}}^{(i)}) \gets \mathcal{E}(D, S^{(i)})\) \COMMENT{Evaluate current summary}
    
    \IF{\(\text{score}_{\text{cov}} \geq T_{\text{cov}}\) AND \(\text{score}_{\text{cons}} \geq T_{\text{cons}}\)}
        \RETURN \(S^{(i)}\) \COMMENT{Quality thresholds satisfied}
    \ENDIF
    
    \IF{\(\text{score}_{\text{cov}} < T_{\text{cov}}\)}
        \STATE \(\tilde{\mathcal{F}}^{(i)} \gets \text{ConstructCoverageFeedback}(\mathcal{F}_{\text{cov}}^{(i)})\) \COMMENT{Natural-language feedback}
    \ELSIF{\(\text{score}_{\text{cons}} < T_{\text{cons}}\)}
        \STATE \(\tilde{\mathcal{F}}^{(i)} \gets \text{ConstructConsistencyFeedback}(\mathcal{F}_{\text{cons}}^{(i)})\)
    \ENDIF
    
    \STATE \(S^{(i+1)} \gets \mathcal{R}(D, S^{(i)}, \tilde{\mathcal{F}}^{(i)})\) \COMMENT{Generate refined summary}
    \STATE \(i \gets i + 1\)
\ENDWHILE

\RETURN \(S^{(i)}\) \COMMENT{Return best summary after max iterations}
\end{algorithmic}
\end{algorithm}

The key innovation lies in converting structured feedback into natural language instructions that explicitly identify missing information or inconsistent facts.

\subsubsection{Coverage-Oriented Feedback}

Coverage feedback identifies missing key information by converting unanswered questions from the evaluation module into natural language instructions (Prompt~\ref{prompt:coverage-feedback}. This approach provides explicit guidance on what content to add without requiring the LLM to independently identify coverage gaps in long documents. By presenting missing information as questions, the feedback suggests what to include while allowing flexibility in phrasing, maintaining natural language flow.

\vspace{6pt}

{\small
\begin{tcolorbox}[
    colback=purple!5, 
    colframe=purple!50!black, 
    boxrule=1pt, 
    rounded corners, 
    title=\textbf{Prompt III-D1: Coverage-Oriented Refinement}, 
    fonttitle=\bfseries,
    label={prompt:coverage-feedback}
]

\textbf{Task} \\
Revise the summary to address key questions from the source document that are not adequately covered.

\vspace{4pt}
\textbf{Context} \\
Your initial summary does not sufficiently cover some important information from the source document.

\vspace{4pt}
\textbf{Requirements}
\begin{itemize}[leftmargin=1.5em, itemsep=2pt, parsep=0pt, topsep=2pt]
    \item Address all the questions listed below in your revised summary
    \item Maintain conciseness and readability
    \item Ensure all added information is accurate and from the source document
    \item Preserve the quality of existing content while adding missing information
\end{itemize}

\vspace{4pt}
\textbf{Input} \\
Document \(D\): \textit{<document\_content>} \\
Initial Summary \(S^{(i)}\): \textit{<current\_summary>} \\
Missing Information \(\mathcal{F}_{\text{cov}}\): \textit{<unanswered\_questions>}

\vspace{4pt}
\textbf{Output Format}
\begin{verbatim}
<revised_summary>
\end{verbatim}

\end{tcolorbox}
}

\subsubsection{Consistency-Oriented Feedback}

Consistency feedback corrects factual errors by converting inconsistent fact triplets into natural language instructions that provide correct facts from the source document (Prompt~\ref{prompt:consistency-feedback}. By providing both the question identifying the claim and the ground truth answer, the feedback enables precise correction without requiring re-extraction from lengthy documents. This fact-level granularity reduces ambiguity and makes corrections more reliable, which is particularly important for high-stakes applications requiring factual accuracy.

\vspace{6pt}

{\small
\begin{tcolorbox}[
    colback=red!5, 
    colframe=red!50!black, 
    boxrule=1pt, 
    rounded corners, 
    title=\textbf{Prompt III-D2: Consistency-Oriented Refinement}, 
    fonttitle=\bfseries,
    label={prompt:consistency-feedback}
]

\textbf{Task} \\
Revise the summary to ensure factual consistency with the source document by correcting inaccurate or inconsistent statements.

\vspace{4pt}
\textbf{Context} \\
Your initial summary contains some facts that do not align with information in the source document.

\vspace{4pt}
\textbf{Requirements}
\begin{itemize}[leftmargin=1.5em, itemsep=2pt, parsep=0pt, topsep=2pt]
    \item Ensure all facts align with the ground truth answers provided below
    \item Correct any inaccurate or inconsistent statements
    \item Maintain the overall structure and readability of the summary
    \item Do not add or remove information beyond what is necessary for consistency
\end{itemize}

\vspace{4pt}
\textbf{Input} \\
Document \(D\): \textit{<document\_content>} \\
Initial Summary \(S^{(i)}\): \textit{<current\_summary>} \\
Ground Truth Facts \(\mathcal{F}_{\text{cons}}\): \textit{<question\_answer\_pairs>}

\vspace{4pt}
\textbf{Ground Truth Facts Format}
\begin{verbatim}
Question: <question_text>
Correct Answer: <ground_truth_answer>
...
\end{verbatim}

\vspace{4pt}
\textbf{Output Format}
\begin{verbatim}
<revised_summary>
\end{verbatim}

\end{tcolorbox}
}

\section{Experiments and Results}
\subsection{Datasets}

\begin{figure}
    \centering
    \includegraphics[width=\linewidth]{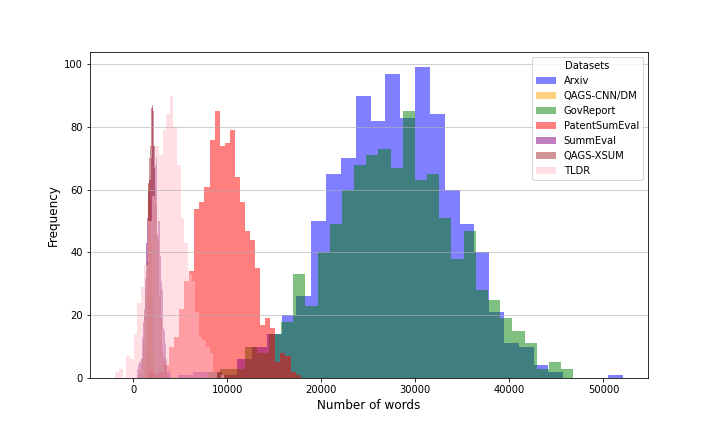}
    \caption{Source document length distributions}
    \label{fig:length-source}
\end{figure}

\begin{figure}
    \centering
    \includegraphics[width=\linewidth]{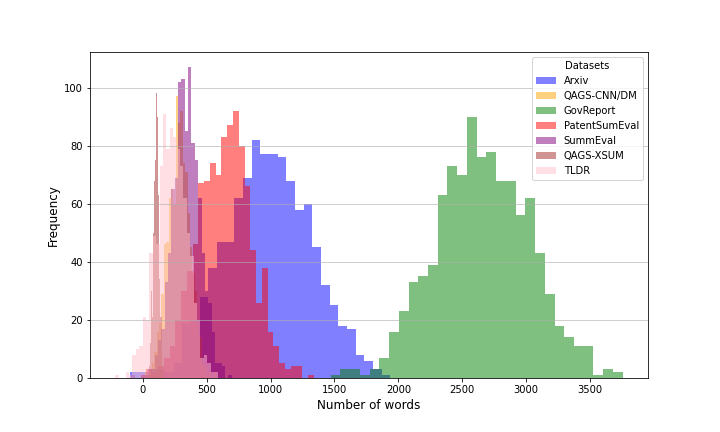}
    \caption{Model-generated summary length distributions}
    \label{fig:length-summary}
\end{figure}

We evaluate our framework on seven human-annotated datasets covering five domains. For evaluation module validation (RQ1-RQ2), we use all seven datasets. For self-refinement module validation (RQ3), we use five datasets with sufficient low-quality summaries: Patent, CNNDM (CNN and DM splits), Arxiv, and PubMed. These datasets differ markedly in both input length and annotation protocol, allowing us to test robustness from short-form summarization to long-document settings. Table~\ref{tab:datasets-overview} provides a compact summary of these datasets. Figure~\ref{fig:length-source} and Figure~\ref{fig:length-summary} further compare the length distributions of source documents and model-generated summaries.

\begin{table*}[t]
\centering
\small
\caption{Summary of datasets used in our experiments. ``Doc'' and ``Sum'' report mean (std.) word counts. ``Human Eval.'' describes the rating scale and aggregation. ``Dims'' lists the annotated quality dimensions. $^{*}$Our proposed dataset. $^{\dagger}$Sentence-level binary evaluations averaged to summary-level scores. $^{\ddagger}$Sentence-level binary consistency judgments aggregated via averaging. Dims abbreviations: Coh = coherence; Flu = fluency; Clar = clarity; Acc = accuracy; Cons = factual consistency; Rel = relevance; Cov = coverage.}
\label{tab:datasets-overview}
\resizebox{2\columnwidth}{!}{%
\begin{tabular}{l l r r r r l l}
\hline
\textbf{Dataset} & \textbf{Domain} & \textbf{Models} & \textbf{Docs} & \textbf{Doc} & \textbf{Sum} & \textbf{Human Eval.} & \textbf{Dims} \\
\hline
SummEval~\cite{fabbri2021summeval} & News & 23 & 100 & 2,160 (601) & 343 (105) & Likert 1--5 & Coh/Flu/Cons/Rel \\
Arxiv~\cite{koh2022far} & Scientific & 12 & 17 & 27,193 (6,558) & 972 (342) & Binary$^{\dagger}$ & Cons/Rel \\
GovReport~\cite{koh2022far} & Government & 12 & 17 & 27,631 (6,943) & 2,658 (363) & Binary$^{\dagger}$ & Cons/Rel \\
TLDR~\cite{stiennon2020learning} & Social media & 8 & 788 & 3,756 (1,866) & 227 (123) & Likert 1--7 & Coh/Acc/Cov/Overall \\
QAGS-XSUM~\cite{wang2020asking} & News & 1 & 239 & 2,084 (514) & 105 (22) & Binary$^{\ddagger}$ & Cons \\
QAGS-CNN/DM~\cite{wang2020asking} & News & 1 & 235 & 1,791 (192) & 284 (83) & Binary$^{\ddagger}$ & Cons \\
PatentSumEval$^{*}$ & Patents & 6 & 30 & 9,754 (2,727) & 647 (209) & Likert 1--5 & Clar/Acc/Cov/Overall \\
\hline
\end{tabular}%
}
\vspace{0.1cm}
\end{table*}

% \begin{figure}[t]
% \centering
% \begin{subfigure}{0.48\textwidth}
%     \centering
%     \includegraphics[width=\textwidth]{Figures/length_srcdoc_allds.png}
%     \caption{Source document length distributions}
%     \label{fig:length-source}
% \end{subfigure}%
% \hspace{0.01\textwidth}%
% \begin{subfigure}{0.48\textwidth}
%     \centering
%     \includegraphics[width=\textwidth]{Figures/length_sum_allds.png}
%     \caption{Model-generated summary length distributions}
%     \label{fig:length-summary}
% \end{subfigure}
% \caption{Length distributions (in words) of source documents and system summaries across the seven datasets.}
% \label{fig:length-distribution}
% \end{figure}

\noindent\textbf{SummEval.} 100 CNN/DailyMail articles with summaries from 23 systems (16 extractive, 7 abstractive). Five crowdworkers and three experts provide Likert 1--5 ratings for coherence, consistency, fluency, and relevance~\cite{fabbri2021summeval}.

\noindent\textbf{Arxiv \& GovReport.} Long-document summarization benchmarks evaluated by three experts using sentence-level binary judgments on factual consistency and relevance, aggregated to summary-level scores~\cite{koh2022far}. Arxiv uses scientific papers (arXivPubMed), while GovReport focuses on government reports.

\noindent\textbf{TLDR.} 788 Reddit posts with summaries from 8 reinforcement-learning variants. Ten annotators rate each summary on a 1--7 Likert scale for coherence, accuracy, coverage, and overall quality~\cite{stiennon2020learning}.

\noindent\textbf{QAGS-XSUM \& QAGS-CNN/DM.} News summarization sets where three crowdworkers annotate sentence-level factual consistency for summaries generated by a single system (BART for XSUM; Bottom-Up for CNN/DailyMail), and labels are averaged to form summary-level consistency scores~\cite{wang2020asking}.

\subsection{PatentSumEval Benchmark Construction}
\label{sec:patenteval}

This section describes how we built PatentSumEval, a human-annotated benchmark for patent summarization evaluation. Our goal is to evaluate summarization methods on legally relevant technical documents, where factual precision and coverage of novelty are critical.

\subsubsection{Data Collection and Inputs}

We collected 30 patents related to communication and streaming technologies from Google Patents\footnote{\url{https://patents.google.com}}. Instead of feeding the full patent text (which is often extremely long and repetitive), we focus on the two parts most informative for summarization and legal interpretation: (i) the abstract, which provides a high-level overview of the invention, and (ii) the claims, which specify the protected scope and key novel elements. For each patent, we concatenate the abstract and claims to form the model input. This differs from prior patent summarization datasets (e.g., BIGPATENT~\cite{sharma2019bigpatent}) that primarily rely on abstracts.

\subsubsection{Summary Generation and Systems}

For each patent, we generate one summary with each of six models, resulting in 180 system summaries. The model set is chosen to cover diverse architectural families, including a domain-adapted patent model (HUPD-T5-base), general-purpose pretrained models (XLNet, BART, Pegasus), a long-context model (LongT5), and an instruction-tuned LLM (GPT-3.5-turbo).

\subsubsection{Human Annotation Protocol}

We recruited three Master's students with computer science/engineering backgrounds to annotate all summaries. Annotations were collected via APPEN\footnote{\url{https://client.appen.com}}, using a 5-point Likert scale (1=Poor, 5=Excellent). To mitigate careless responses, we inserted quality-control items throughout the task and removed inconsistent submissions.

Annotators rated each summary on two dimensions:
\begin{itemize}[leftmargin=1.5em, itemsep=2pt, parsep=0pt, topsep=2pt]
    \item \textit{Factual Consistency (Accuracy)}: whether statements in the summary are supported by the source document without fabrication or hallucination.
    \item \textit{Coverage}: how comprehensively the summary captures key invention details and patent claims from the source.
\end{itemize}

\subsection{Evaluation Metrics}

\paragraph{\textbf{Conventional QA-based metrics.}} We compare against representative QA-based baselines including QAEval~\cite{deutsch2021towards}, QuestEval~\cite{scialom2021questeval}, and SummaQA~\cite{scialom2019answers}. These methods convert evaluation into a question answering problem and typically aggregate answer matching scores into a single scalar.

\paragraph{\textbf{LLM-based QA evaluation (\(\textsc{llmqa}\)).}} We evaluate our \(\textsc{llmqa}\) metrics with two LLM backends. We compute the coverage score by asking whether the summary can answer salient questions extracted from the source document:
\begin{equation}
\textsc{llmqa}_{\text{cov}} = \frac{n_a}{|Q_d|}.
\end{equation}
where $n_a$ is the number of questions in $Q_d$ that are answerable from the summary.
For factual consistency, we generate question--answer pairs from the summary and verify their answers against the source using a similarity score $s_i = \text{Sim}(a_i^s, a_i^d)$ with threshold $\tau$:
\begin{equation}
\textsc{llmqa}_{\text{cons}} = \frac{1}{|Q_s|} \sum_{i=1}^{|Q_s|} \mathbb{I}(s_i > \tau) \cdot s_i.
\end{equation}
We instantiate $\text{Sim}(\cdot)$ using exact/partial match, ROUGE-1 F1, or cosine similarity between answer embeddings.

\subsection{Implementation Details and Evaluation Protocol}

We implement our QA-based evaluation module using two open-source LLMs: Llama-3.1-8B~\cite{grattafiori2024llama} and Linkbricks-V6-32B\footnote{\url{https://huggingface.co/Saxo/Linkbricks-Horizon-AI-Avengers-V6-32B}}. Both models are used for question generation, answer extraction, and consistency verification. The temperature hyperparameter is set to $10^{-10}$ to ensure deterministic and stable evaluation results.

For coverage evaluation, we generate 6--12 questions from source documents. For factual consistency evaluation, we generate 4--10 question-answer pairs from summaries. We evaluate three answer similarity measures: exact/partial match (empm), ROUGE-1 F1 (rouge)~\cite{chin2004rouge}, and cosine similarity (cossim) using sentence embeddings. A threshold $\tau = 0.6$ is applied to distinguish consistent from inconsistent facts.

We assess metric quality using Kendall's $\tau_b$ rank correlation with human judgments. Statistical significance is tested via permutation tests at $p < 0.05$ (*), $p < 0.01$ (**), and $p < 0.001$ (***). We compare against three QA-based baselines: QAEval~\cite{deutsch2021towards}, QuestEval~\cite{scialom2021questeval}, and SummaQA~\cite{scialom2019answers}.

\subsection{Experimental Results}

Our experimental evaluation addresses the three research questions formulated 
in Section~\ref{intro}:

\begin{itemize}
    \item \textbf{RQ1 \& RQ2}:
          We evaluate the \emph{QA-based evaluation module} by measuring its 
          correlation with human judgments across seven datasets and analyzing 
          design choices.
    \item \textbf{RQ3}: 
          We evaluate the \emph{self-refinement module} by measuring quality 
          improvements on five datasets using feedback from the evaluation module.
    \item \textbf{Human validation}: 
          We validate the quality of questions and answers generated by the 
          evaluation module.
\end{itemize}

\subsubsection{QA-Based Evaluation Performance}
\label{4.5.1}

\begin{table*}[htbp]
\centering
\caption{Correlation of QA-based evaluation metrics with human judgments across seven datasets. Our QA-based evaluation module (LongSumEval) is compared against baseline QA metrics (QAEval, QuestEval, SummaQA). Kendall's $\tau_b$ correlation coefficients are reported. Significance levels: * $p < 0.05$, ** $p < 0.01$, *** $p < 0.001$. Answer similarity measures for consistency evaluation: empm = exact/partial match; rouge = ROUGE-1 F1; cossim = cosine similarity. -- indicates not applicable.}
\label{tab:overall-performance}
\resizebox{2\columnwidth}{!}{%
\begin{tabular}{llc|ccccccc}
\hline
\textbf{Method} & \textbf{Dimension} & \textbf{Sim} & \textbf{SummEval} & \textbf{QAGS-C} & \textbf{QAGS-X} & \textbf{Arxiv} & \textbf{GovRe} & \textbf{TLDR} & \textbf{Patent} \\ 
\hline
\multicolumn{10}{l}{\textit{Baseline: QA-based Metrics}} \\
\multirow{3}{*}{QAEval} & Consistency & empm & 0.000 & 0.000 & 0.000 & 0.000 & 0.000 & -0.015 & 0.000 \\
 & Consistency & ans & 0.417* & -0.026 & -0.129* & 0.030 & 0.303 & 0.046*** & -0.200 \\
 & Coverage & -- & 0.000 & -- & -- & 0.000 & 0.000 & 0.053*** & 0.000 \\
\cline{2-10}
\multirow{2}{*}{QuestEval} & Consistency & -- & 0.567** & 0.173*** & 0.182*** & 0.273 & -0.182 & 0.245*** & 0.000 \\
 & Coverage & -- & 0.200 & -- & -- & -0.273 & -0.364 & 0.161*** & 0.105 \\
\cline{2-10}
\multirow{2}{*}{SummaQA} & Consistency & prob & 0.500** & 0.156*** & 0.085 & 0.333 & -0.091 & 0.139*** & 0.600 \\
 & Coverage & -- & 0.500** & -- & -- & 0.273 & -0.030 & 0.219*** & 0.527 \\
\hline
\multicolumn{10}{l}{\textit{LongSumEval (QA-based Evaluation Module)}} \\
\multirow{4}{*}{\begin{tabular}[c]{@{}l@{}}LongSumEval\\ (Llama-3.1-8B)\end{tabular}} & Consistency & empm & 0.483*** & 0.278*** & 0.102 & 0.333 & -0.091 & 0.195*** & 0.000 \\
 & Consistency & rouge & 0.450*** & 0.296*** & 0.082 & 0.303 & -0.061 & 0.199*** & 0.000 \\
 & Consistency & cossim & 0.367** & 0.183* & 0.072 & 0.303* & -0.333 & 0.144*** & 0.200 \\
 & Coverage & -- & 0.350 & -- & -- & 0.424* & 0.242 & 0.094*** & 0.527* \\
\cline{2-10}
\multirow{4}{*}{\begin{tabular}[c]{@{}l@{}}LongSumEval\\ (Linkbricks-V6-32B)\end{tabular}} & Consistency & empm & 0.683*** & 0.293*** & 0.203*** & 0.485* & 0.030 & 0.245*** & 0.800 \\
 & Consistency & rouge & 0.667*** & 0.311*** & 0.191*** & 0.455** & 0.121 & 0.252*** & 0.800 \\
 & Consistency & cossim & 0.333** & 0.181* & 0.002 & 0.182 & 0.273 & 0.119*** & 0.800 \\
 & Coverage & -- & 0.500** & -- & -- & 0.443 & 0.515* & 0.262*** & 0.738* \\
\hline
\end{tabular}%
}
\end{table*}

Table~\ref{tab:overall-performance} presents the correlation between automatic evaluation metrics and human judgments across all seven datasets using Kendall's $\tau_b$ coefficient. 

\textbf{Superior consistency evaluation.} Our QA-based evaluation module with Linkbricks-V6-32B achieves the best performance on five out of seven datasets for consistency evaluation. On SummEval, it attains the highest correlation ($\tau_b = 0.683$, $p < 0.001$ with empm; $\tau_b = 0.667$, $p < 0.001$ with rouge), substantially outperforming QuestEval ($\tau_b = 0.567$) and SummaQA ($\tau_b = 0.500$). For QAGS-XSUM, our QA-based evaluation module achieves $\tau_b = 0.203$ ($p < 0.001$) with empm and $\tau_b = 0.191$ ($p < 0.001$) with rouge, both surpassing baseline methods. On the scientific writing dataset (Arxiv), it demonstrates strong performance with $\tau_b = 0.485$ ($p < 0.05$) using empm and $\tau_b = 0.455$ ($p < 0.01$) using rouge. The PatentSumEval dataset shows particularly strong results, with our QA-based evaluation module achieving $\tau_b = 0.800$ for consistency evaluation across all similarity measures.

\textbf{Competitive coverage performance.} Our QA-based evaluation module with Linkbricks-V6-32B attains second-best results on SummEval ($\tau_b = 0.500$, $p < 0.01$) and strong correlations on GovReport ($\tau_b = 0.515$, $p < 0.05$), TLDR ($\tau_b = 0.262$, $p < 0.001$), and PatentSumEval ($\tau_b = 0.738$, $p < 0.05$). Notably, our evaluation module with Llama-3.1-8B achieves the best coverage performance on Arxiv ($\tau_b = 0.424$, $p < 0.05$), demonstrating robustness across different LLM backends.

\textbf{Impact of similarity measures.} ROUGE-1 F1 generally provides the most stable correlations across datasets, achieving best or second-best performance on five datasets (SummEval, QAGS-CNNDM, Arxiv, TLDR). Exact/partial match performs competitively on news datasets (SummEval, QAGS-XSUM) but shows lower correlations on scientific writing. Cosine similarity exhibits more variable performance, suggesting that lexical overlap measures are more reliable than semantic similarity for answer matching in summarization evaluation.

\textbf{Baseline limitations.} Among the QA-based baselines, performance varies considerably. QAEval demonstrates the weakest results, with correlations at or near zero across most datasets and dimensions, except for modest performance on SummEval consistency ($\tau_b = 0.417$, $p < 0.05$) using answer similarity. This suggests that QAEval's exact-match approach and question generation strategy may not capture the nuances required for diverse summarization evaluation. SummaQA shows more competitive results, achieving $\tau_b = 0.500$ ($p < 0.01$) on SummEval for both consistency and coverage, and strong performance on PatentSumEval ($\tau_b = 0.600$ for consistency, $\tau_b = 0.527$ for coverage). However, it exhibits weaker correlations on QAGS datasets and mixed results on long documents. QuestEval achieves competitive results on SummEval for consistency ($\tau_b = 0.567$, $p < 0.01$) but performs poorly or even negatively on long-document datasets (Arxiv, GovReport), with $\tau_b = -0.182$ on GovReport consistency and $\tau_b = -0.364$ on GovReport coverage. This suggests that conventional QA-based metrics designed for short-form summarization may struggle with lengthy inputs where question generation and answer extraction become more challenging. In contrast, our QA-based evaluation module maintains robust performance across both short-form and long-document settings, demonstrating the effectiveness of LLM-based question-answering for diverse evaluation scenarios.

\subsubsection{Self-Refinement Effectiveness}
\label{4.5.2}

\begin{table*}[htbp]
\centering
\caption{Self-refinement performance across five datasets. Scores reported as mean (standard deviation). IS = initial summary; RS = refined summary; $\Delta$ = relative improvement (\%). Thresholds: low coverage $\text{score}_{\text{cov}} < 0.60$, low consistency $\text{score}_{\text{cons}} < 0.73$. Metrics computed using Linkbricks-V6-32B with ROUGE-1 F1 answer similarity.}
\label{tab:refinement-performance}
\resizebox{2\columnwidth}{!}{%
\begin{tabular}{ll|ccc|ccc|ccc}
\hline
\multirow{2}{*}{\textbf{Dataset}} & \multirow{2}{*}{\textbf{Dimension}} & 
  \multicolumn{3}{c|}{\textbf{All Summaries}} & 
  \multicolumn{3}{c|}{\textbf{Low Coverage ($\text{score}_{\text{cov}} < 0.60$)}} & 
  \multicolumn{3}{c}{\textbf{Low Consistency ($\text{score}_{\text{cons}} < 0.73$)}} \\
\cline{3-11}
 & & \textbf{IS} & \textbf{RS} & \textbf{$\Delta$} & \textbf{IS} & \textbf{RS} & \textbf{$\Delta$} & \textbf{IS} & \textbf{RS} & \textbf{$\Delta$} \\ 
\hline
\multirow{2}{*}{Patent} 
  & Coverage & 0.549 (0.186) & 0.767 (0.141) & \textbf{+39.71} & 0.430 (0.122) & 0.790 (0.151) & \textbf{+83.72} & 0.514 (0.187) & 0.806 (0.168) & \textbf{+56.81} \\
  & Consistency & 0.820 (0.158) & 0.875 (0.127) & \textbf{+6.71} & 0.806 (0.165) & 0.872 (0.137) & \textbf{+8.19} & 0.582 (0.099) & 0.858 (0.142) & \textbf{+47.42} \\ 
\hline
\multirow{2}{*}{CNN/DM (CNN)} 
  & Coverage & 0.632 (0.192) & 0.737 (0.161) & \textbf{+16.61} & 0.445 (0.108) & 0.740 (0.168) & \textbf{+66.29} & 0.634 (0.194) & 0.720 (0.183) & \textbf{+13.56} \\
  & Consistency & 0.757 (0.191) & 0.807 (0.171) & \textbf{+6.61} & 0.764 (0.192) & 0.776 (0.181) & \textbf{+1.57} & 0.561 (0.134) & 0.762 (0.194) & \textbf{+35.83} \\ 
\hline
\multirow{2}{*}{CNN/DM (DM)} 
  & Coverage & 0.644 (0.181) & 0.701 (0.177) & \textbf{+8.85} & 0.459 (0.096) & 0.752 (0.162) & \textbf{+63.83} & 0.645 (0.176) & 0.743 (0.170) & \textbf{+15.19} \\
  & Consistency & 0.785 (0.185) & 0.786 (0.185) & \textbf{+0.13} & 0.788 (0.186) & 0.790 (0.186) & \textbf{+0.25} & 0.566 (0.130) & 0.766 (0.190) & \textbf{+35.34} \\ 
\hline
\multirow{2}{*}{Arxiv} 
  & Coverage & 0.532 (0.231) & 0.586 (0.279) & \textbf{+10.15} & 0.366 (0.152) & 0.604 (0.306) & \textbf{+65.03} & 0.467 (0.249) & 0.586 (0.313) & \textbf{+25.48} \\
  & Consistency & 0.778 (0.205) & 0.732 (0.241) & -5.91 & 0.748 (0.223) & 0.708 (0.255) & -5.35 & 0.538 (0.164) & 0.691 (0.267) & \textbf{+28.44} \\ 
\hline
\multirow{2}{*}{PubMed} 
  & Coverage & 0.535 (0.192) & 0.725 (0.171) & \textbf{+35.51} & 0.415 (0.128) & 0.727 (0.188) & \textbf{+75.18} & 0.524 (0.196) & 0.728 (0.207) & \textbf{+38.93} \\
  & Consistency & 0.762 (0.213) & 0.796 (0.196) & \textbf{+4.46} & 0.753 (0.217) & 0.768 (0.209) & \textbf{+1.99} & 0.518 (0.165) & 0.752 (0.224) & \textbf{+45.17} \\ 
\hline
\end{tabular}%
}
\end{table*}

To validate the actionability of our QA-based evaluation module's structured feedback, we examine its effectiveness in guiding the self-refinement module for iterative quality improvement. We evaluate performance across five datasets under three conditions to assess sensitivity to initial quality: all summaries establish baseline effectiveness, while subsets with low coverage ($\text{score}_{\text{cov}} < 0.60$) or low consistency ($\text{score}_{\text{cons}} < 0.73$) test whether targeted feedback achieves stronger improvements when specific deficiencies are present. Results are reported in Table~\ref{tab:refinement-performance}.

\textbf{Substantial improvements on low-quality summaries.} The self-refinement module demonstrates its strongest impact when applied to summaries with specific quality deficiencies. For low-coverage summaries, coverage scores improve dramatically across all datasets: Patent (+83.72\%), CNN/DM-CNN (+66.29\%), CNN/DM-DM (+63.83\%), Arxiv (+65.03\%), and PubMed (+75.18\%). Similarly, for low-consistency summaries, consistency scores show marked improvements: Patent (+47.42\%), CNN/DM-CNN (+35.83\%), CNN/DM-DM (+35.34\%), Arxiv (+28.44\%), and PubMed (+45.17\%). These results validate the effectiveness of targeted feedback-guided refinement for addressing specific quality deficiencies.

\textbf{Consistent coverage gains across all datasets.} When applied to all summaries regardless of initial quality, the refinement module consistently improves coverage scores. Patent summaries show the largest overall coverage improvement (+39.71\%), followed by PubMed (+35.51\%), CNN/DM-CNN (+16.61\%), Arxiv (+10.15\%), and CNN/DM-DM (+8.85\%). The particularly strong performance on Patent and PubMed datasets suggests that the refinement approach is especially effective for technical and scientific domains where coverage of key technical details is critical.

\textbf{Moderate consistency improvements with trade-offs.} Consistency refinement shows more modest gains when applied to all summaries, ranging from +0.13\% (CNN/DM-DM) to +6.71\% (Patent). Notably, Arxiv exhibits a slight decrease in consistency scores for all summaries (-5.91\%) and low-coverage summaries (-5.35\%), suggesting a potential trade-off between coverage and consistency during refinement. However, when specifically targeting low-consistency summaries, all datasets show positive improvements, with the most substantial gains on Patent (+47.42\%) and PubMed (+45.17\%).

\textbf{Domain-dependent refinement effectiveness.} Patent and PubMed datasets consistently demonstrate the strongest refinement gains across both dimensions, with coverage improvements exceeding 35\% and consistency improvements reaching 45\% for low-quality summaries. In contrast, news datasets (CNN/DM) show more modest improvements, particularly for consistency refinement on already high-quality summaries. This pattern suggests that the structured, fact-dense nature of technical documents provides more opportunities for targeted refinement, while news summaries may already capture salient information effectively in their initial generation.

\subsubsection{Human Validation of Question-Answer Quality}
\label{4.5.3}

\begin{table*}[htbp]
\centering
\caption{Human evaluation of generated question and answer quality. Three evaluators independently assessed QA pairs generated by Linkbricks-V6-32B. Question evaluation: 90 responses (450 questions total). Answer evaluation: 78 responses (390 answers total). Inter-rater agreement: Krippendorff's $\alpha = 0.858$ (questions), $\alpha = 0.705$ (answers).}
\label{tab:human-eval-qa}
\resizebox{1.9\columnwidth}{!}{%
\begin{tabular}{lllrr}
\hline
\multicolumn{3}{l}{\textbf{Evaluation Criterion}} & \textbf{Yes (\%)} & \textbf{No (\%)} \\ 
\hline
\multicolumn{5}{l}{\textit{Question Quality Assessment}} \\
\multicolumn{3}{l}{\quad Does the question ask for key information from the document?} & 91.67 & 8.33 \\
\multicolumn{3}{l}{\quad Can the answer be explicitly stated or logically inferred from the document?} & 98.00 & 2.00 \\
\multicolumn{3}{l}{\quad Do the questions cover all key aspects that a summary should address?} & 60.00 & 40.00 \\ 
\hline
\multicolumn{5}{l}{\textit{Answer Quality Assessment}} \\
\multicolumn{3}{l}{\quad Is the answer relevant to the question?} & 100.00 & 0.00 \\
 & \multicolumn{2}{l}{\quad\quad - If yes: Does the answer state ``no answer''?} & 24.62 & 75.38 \\
 &  & \quad\quad\quad + If yes: Is ``no answer'' justified (missing info)? & 87.50 & 12.50 \\
 &  & \quad\quad\quad + If no: Is the answer factually consistent? & 99.00 & 1.00 \\ 
\hline
\end{tabular}%
}
\end{table*}

To assess the quality of our QA-based evaluation module, we conduct human 
evaluation of the generated questions and answers. Table~\ref{tab:human-eval-qa} presents the results of a human evaluation study assessing the quality of questions and answers generated by Linkbricks-V6-32B. Three independent evaluators assessed 450 questions and 390 answers, achieving strong inter-rater agreement (Krippendorff's $\alpha = 0.858$ for questions, $\alpha = 0.705$ for answers).

\textbf{High-quality question generation.} The vast majority of generated questions focus on key document information, with 91.67\% asking for salient content rather than trivial details. Furthermore, 98.00\% of questions can be explicitly answered or logically inferred from the source document, demonstrating that the LLM generates well-grounded questions that are answerable from the provided context. This high rate of answerability validates the approach of using LLM-generated questions as a basis for evaluation, as they represent verifiable factual content rather than speculative or unanswerable queries.

\textbf{Limited coverage comprehensiveness.} While individual questions demonstrate high quality, evaluators judged that only 60.00\% of question sets comprehensively cover all key aspects that a summary should address. This suggests that while the generated questions are relevant and answerable, the current question generation approach may miss some important dimensions of document content. This finding highlights a potential area for improvement in the question generation process, such as increasing question diversity or implementing more structured coverage of different document sections.

\textbf{Perfect answer relevance and high factual consistency.} All extracted answers (100.00\%) are relevant to their corresponding questions, indicating reliable question-answer alignment. Among answers that provide substantive content (75.38\%), 99.00\% are factually consistent with the source document, with only 1.00\% containing factual errors. This near-perfect consistency rate demonstrates the LLM's strong capability for accurate information extraction from source documents.

\textbf{Appropriate handling of unanswerable questions.} The system correctly identifies when information is missing, with 24.62\% of answers appropriately returning "no answer" or equivalent responses. Of these cases, 87.50\% are justified by genuinely missing information in the source, while only 12.50\% represent false negatives where the information was actually present. This high precision in detecting unanswerable questions validates the coverage assessment methodology, which relies on distinguishing between answerable and unanswerable questions to measure summary completeness.

\subsubsection{Sensitivity Analysis}

\begin{table}[htbp]
\centering
\caption{Impact of question quantity on coverage evaluation performance. Kendall's Tau-b correlation with human judgments. $|Q_d|$ = number of questions generated from source document. Significance levels: * $p < 0.05$, ** $p < 0.01$, *** $p < 0.001$. Best results per dataset are \textbf{bold}, second-best are \underline{underlined}.}
\label{tab:sensitivity-question-number-cov}
\begin{tabular}{ccc}
\hline
\textbf{$|Q_d|$ Range} & \textbf{Arxiv} & \textbf{SummEval} \\ 
\hline
3--6   & \textbf{0.667***} & 0.383* \\
6--12  & 0.443 & \underline{0.500**} \\
9--15  & 0.394 & \textbf{0.517**} \\
12--18 & \underline{0.576*} & 0.417* \\ 
\hline
\end{tabular}
\end{table}

\begin{table*}[htbp]
\centering
\caption{Impact of question quantity on factual consistency evaluation performance. Kendall's Tau-b correlation with human judgments. $|Q_s|$ = number of questions generated from summary. Answer similarity measures: empm = exact/partial match; rouge = ROUGE-1 F1; cossim = cosine similarity. Significance levels: * $p < 0.05$, ** $p < 0.01$, *** $p < 0.001$. Best results per dataset and similarity measure are \textbf{bold}, second-best are \underline{underlined}.}
\label{tab:sensitivity-question-number-cons}
\resizebox{0.7\textwidth}{!}{%
\begin{tabular}{c|ccc|ccc}
\hline
& \multicolumn{3}{c|}{\textbf{Arxiv}} & \multicolumn{3}{c}{\textbf{SummEval}} \\ 
\hline
\textbf{$|Q_s|$ Range} & \textbf{empm} & \textbf{rouge} & \textbf{cossim} & \textbf{empm} & \textbf{rouge} & \textbf{cossim} \\ 
\hline
3--5   & 0.121 & 0.303 & \textbf{0.576*} & 0.667*** & 0.667*** & 0.533*** \\
5--7   & \underline{0.424*} & \textbf{0.455*} & \underline{0.412} & \textbf{0.728***} & \underline{0.700***} & \textbf{0.633***} \\
7--12  & 0.303 & 0.303 & 0.137 & \underline{0.700***} & \textbf{0.733***} & \underline{0.517***} \\ 
\hline
\end{tabular}%
}
\end{table*}

\begin{table}[htbp]
\centering
\caption{Impact of similarity threshold on factual consistency evaluation performance. Kendall's Tau-b correlation with human judgments. Answer similarity measures: empm = exact/partial match; rouge = ROUGE-1 F1; cossim = cosine similarity. Optimal question range: $|Q_s|$ = 5--7 based on Table~\ref{tab:sensitivity-question-number-cons}. Significance levels: * $p < 0.05$, ** $p < 0.01$, *** $p < 0.001$. Best results per dataset and similarity measure are \textbf{bold}, second-best are \underline{underlined}.}
\label{tab:sensitivity-threshold}
\resizebox{0.95\columnwidth}{!}{%
\begin{tabular}{c|ccc|ccc}
\hline
& \multicolumn{3}{c|}{\textbf{Arxiv}} & \multicolumn{3}{c}{\textbf{SummEval}} \\ 
\hline
\textbf{$\tau$} & \textbf{empm} & \textbf{rouge} & \textbf{cossim} & \textbf{empm} & \textbf{rouge} & \textbf{cossim} \\ 
\hline
0.2 & \textbf{0.485*} & \underline{0.545**} & 0.394* & 0.711*** & 0.667*** & \underline{0.650***} \\
0.3 & \textbf{0.485*} & \textbf{0.606**} & 0.333* & 0.717*** & 0.667*** & \underline{0.650***} \\
0.4 & \textbf{0.485*} & 0.515* & 0.394* & \textbf{0.733***} & 0.667*** & 0.633*** \\
0.5 & \textbf{0.485*} & 0.515* & \textbf{0.424*} & \textbf{0.733***} & \underline{0.717***} & \underline{0.650***} \\
0.6 & \underline{0.424*} & 0.455** & \underline{0.412*} & \underline{0.728***} & 0.700*** & \underline{0.650***} \\
0.7 & \underline{0.424*} & 0.485* & \underline{0.412} & 0.700*** & \textbf{0.733***} & \textbf{0.667***} \\ 
\hline
\end{tabular}%
}
\end{table}

We analyze the sensitivity of our QA-based evaluation module to two key hyperparameters: the number of questions generated and the similarity threshold for consistency evaluation. Tables~\ref{tab:sensitivity-question-number-cov}, \ref{tab:sensitivity-question-number-cons}, and \ref{tab:sensitivity-threshold} present results on Arxiv and SummEval datasets.

\textbf{Dataset-dependent optimal question quantity for coverage.} The optimal number of questions for coverage evaluation varies by dataset characteristics (Table~\ref{tab:sensitivity-question-number-cov}). For Arxiv, a scientific dataset with long documents (mean 27,193 words), fewer questions (3--6) achieve the best correlation ($\tau_b = 0.667$, $p < 0.001$), though a larger range (12--18) also performs well ($\tau_b = 0.576$, $p < 0.05$). In contrast, SummEval, with shorter news articles (mean 2,160 words), benefits from more questions, achieving peak performance at 9--15 questions ($\tau_b = 0.517$, $p < 0.01$). This suggests that longer documents may require either highly selective key questions or more comprehensive coverage, while shorter documents benefit from moderate question quantities.

\textbf{Consistent optimal range for consistency evaluation.} For factual consistency assessment, the 5--7 question range consistently performs best or second-best across both datasets and similarity measures (Table~\ref{tab:sensitivity-question-number-cons}). On SummEval, this range achieves the highest correlations for empm ($\tau_b = 0.728$, $p < 0.001$) and cossim ($\tau_b = 0.633$, $p < 0.001$), and second-best for rouge ($\tau_b = 0.700$, $p < 0.001$). On Arxiv, it yields the best performance for rouge ($\tau_b = 0.455$, $p < 0.05$) and second-best for empm ($\tau_b = 0.424$, $p < 0.05$). This suggests that 5--7 factual claims provide sufficient granularity for consistency verification without introducing noise from excessive question generation.

\textbf{Robust performance across threshold values.} The similarity threshold $\tau$ exhibits relatively stable performance across a wide range of values (Table~\ref{tab:sensitivity-threshold}). For Arxiv with rouge similarity, performance peaks at $\tau = 0.3$ ($\tau_b = 0.606$, $p < 0.01$), while for SummEval with empm, optimal thresholds are $\tau = 0.4$ and $\tau = 0.5$ (both $\tau_b = 0.733$, $p < 0.001$). Notably, correlations remain strong across the range $\tau = 0.2$ to $\tau = 0.7$, with most values achieving statistical significance. This robustness suggests that the evaluation framework is not overly sensitive to threshold selection, making it practical for deployment without extensive hyperparameter tuning.

\textbf{Interaction between similarity measure and threshold.} Different similarity measures exhibit distinct threshold sensitivity patterns. Exact/partial match (empm) shows stable performance across all thresholds on Arxiv (consistently $\tau_b = 0.485$ for $\tau = 0.2$ to $\tau = 0.5$), while ROUGE-1 F1 demonstrates more variation, peaking at lower thresholds ($\tau = 0.3$ on Arxiv). Cosine similarity performs best at moderate thresholds ($\tau = 0.5$ on Arxiv, $\tau = 0.7$ on SummEval). These patterns suggest that lexical measures (empm, rouge) benefit from lower thresholds that capture partial matches, while semantic similarity (cossim) requires higher thresholds to filter low-confidence alignments.

\section{Discussion}

\subsection{Bridging Evaluation and Generation Through Structured Feedback}

Traditional evaluation metrics provide numerical assessments without actionable guidance for improvement. Our framework resolves this disconnect by operationalizing evaluation as structured question-answering feedback that simultaneously quantifies quality and generates executable refinement instructions. The structured approach substantially outperforms generic self-refinement methods~\cite{madaan2024self}, as recent work demonstrates that fine-grained feedback enables more effective refinement than abstract critiques~\cite{xu-etal-2024-llmrefine,wadhwa-etal}. By extracting corrective information during evaluation and presenting it as question-answer pairs, our feedback eliminates the search problem inherent in vague critiques. This design aligns with findings that natural language feedback better matches LLM capabilities than numerical rewards~\cite{yang2025lighthouse}.

The structured feedback achieves decomposition wherein each unanswered question specifies missing information with explicit answers, and each fact triplet identifies errors with ground-truth replacements. This granularity explains differential effectiveness across domains: technical documents with distributed information benefit substantially more from explicit guidance than shorter news articles. The evaluation-as-instruction paradigm extends beyond summarization to any controllable generation task requiring verifiable accuracy~\cite{yu-etal-2025-frame}. Unlike black-box learned metrics, question-answer evaluation generates human-inspectable artifacts enabling domain expert verification~\cite{yang2025lighthouse}, establishing a template for responsible deployment in consequential domains.

\subsection{The Coverage-Consistency Trade-off}

Universal refinement improved coverage while degrading consistency on long documents, whereas selective refinement targeting low-consistency summaries improved both dimensions. This demonstrates that refinement strategy selection matters as critically as the mechanism itself, consistent with recent work identifying tensions between completeness and precision~\cite{samarinas-etal-2025-beyond}. The trade-off arises from retrieval precision-recall dynamics: expanding coverage incentivizes recall maximization, which typically compromises precision. LLMs may retrieve tangentially related content that is technically present but contextually inappropriate, a phenomenon intensifying in long documents where increased search space elevates retrieval error probability. Selective refinement outcomes suggest consistency deficiencies often stem from retrieval errors rather than fabrication, while high-consistency summaries achieved reliability through conservative information inclusion~\cite{yang-etal-2025-confidence}. Three design principles emerge for deployment. First, threshold-based triggering where refinement activates only when scores fall below domain-calibrated thresholds. Second, sequential refinement prioritizing consistency before coverage, as the relationship is asymmetric. Third, dimension-aware targeting wherein refinement strategy adapts to initial quality profiles.

\subsection{Limitations and Future Directions}

Our framework exhibits two considerations for future enhancement. First, while we mitigate LLM evaluation bias through diverse model architectures and human-inspectable question-answer pairs, the LLM-as-judge paradigm warrants continued investigation. Second, our exploration focused on initial refinement iterations, leaving opportunities to characterize extended refinement dynamics and optimize stopping criteria across quality dimensions.

Future work should pursue several promising directions. Multi-evaluator consensus approaches aggregating judgments across diverse models could further strengthen evaluation reliability. Systematic investigation of refinement trajectories across multiple iterations would enable adaptive termination policies that maximize quality gains. Extending dimension-aware strategies through reinforcement learning policies that optimize based on outcome feedback represents a natural progression. Additionally, comparative analysis of LLM versus human revision patterns could reveal generalizable principles for iterative improvement. These directions build upon our framework's foundation of structured feedback and transparent evaluation to advance controllable generation in high-stakes applications.

\section{Conclusion}
We introduce LongSumEval, a framework that bridges evaluation and generation through question-answering feedback producing quantitative scores and structured instructions identifying coverage gaps and factual inconsistencies. Meta-evaluation of our QA-based evaluation module across multiple benchmarks demonstrates superior correlation with human judgments compared to conventional metrics. We show that diverse question types, ROUGE-1 F1 similarity measures, and advanced LLM backends optimize reliability. Our structured feedback enables self-refinement without retraining, substantially improving summary quality. By converting assessment results into actionable instructions specifying unanswered questions and inconsistent fact triplets, we establish that evaluation feedback can serve as executable generation guidance. Our framework generalizes across domains and document lengths. To support reproducibility and real-world adoption, we release the PatentSumEval benchmark and accompanying code for high-stakes applications requiring verifiable accuracy.

\bibliographystyle{IEEEtran}
\bibliography{reference}

\end{document}